\documentclass[letterpaper, 10 pt, conference]{ieeeconf}  

\IEEEoverridecommandlockouts                              

\AtEndDocument{\par\leavevmode} 

\PassOptionsToPackage{hyphens}{url}\usepackage{hyperref}

\usepackage[draft]{pgf}

\usepackage{balance}

\usepackage{graphics} 
\usepackage{graphicx} 
\usepackage{epsfig} 
\usepackage{amsmath, bm} 
\usepackage{amssymb}  
\usepackage{upgreek}

\usepackage[caption=false, font=footnotesize]{subfig}

\usepackage{cuted}
\usepackage{booktabs}
\usepackage{multicol}
\usepackage{multirow}
\usepackage{stfloats}
\usepackage{lipsum}  
\usepackage{breqn}
\usepackage{comment} 

\title{\LARGE \bf
Rough-Terrain Locomotion and Unilateral Contact Force Regulations With a Multi-Modal Legged Robot
}

\author{Kaier Liang, Eric Sihite, Pravin Dangol, Andrew Lessieur, and Alireza Ramezani$^{1}$
\thanks{$^{1}$SiliconSynapse Laboratory, ECE Department, Northeastern University, Boston, MA, USA. emails: \{liang.k, e.sihite, dangol.p, lessieur.a, a.ramezani\} @northeastern.edu}%
}

\begin{document}

\maketitle

\global\csname @topnum\endcsname 0
\global\csname @botnum\endcsname 0

\thispagestyle{empty}
\pagestyle{empty}

\begin{abstract}

Despite many accomplishments by legged robot designers, state-of-the-art bipedal robots are prone to falling over, cannot negotiate extremely rough terrains and cannot directly regulate unilateral contact forces. Our objective is to integrate merits of legged and aerial robots in a single platform. We will show that the thrusters in a bipedal legged robot called \textit{Harpy} can be leveraged to stabilize the robot's frontal dynamics and permit jumping over large obstacles which is an unusual capability not reported before. In addition, we will capitalize on the thrusters action in Harpy and will show that one can avoid using costly optimization-based schemes by directly regulating contact forces using an Reference Governor (RGs). We will resolve gait parameters and re-plan them during gait cycles by only assuming well-tuned supervisory controllers. Then, we will focus on RG-based fine-tuning of the joints desired trajectories to satisfy unilateral contact force constraints.

\end{abstract}


\section{Introduction}
\label{sec:introduction}

Raibert's hopping robots \cite{raibert1984experiments} and Boston Dynamics' robots \cite{raibert2008bigdog} are amongst the most successful examples of legged robots, as they can hop or trot robustly even in the presence of significant unplanned disturbances. Other than these successful examples, a large number of underactuated and fully actuated bipedal robots have also been introduced. Agility Robotics' Cassie \cite{gong2019feedback}, Honda's ASIMO \cite{hirose2006honda} and Samsung's Mahru III \cite{kwon2007biped} are capable of walking, running, dancing and going up and down stairs, and the Yobotics-IHMC \cite{5354430} biped can recover from pushes.
%
Despite these accomplishments, all of these systems are prone to falling over and cannot negotiate extremely rough terrains. Even humans, known for their natural, dynamic and robust gaits cannot recover from severe rough terrain perturbations, external pushes or slippage on icy surfaces. Our goal is to enhance the robustness of these systems through a distributed array of thrusters and nonlinear control. 

\begin{figure}[t]
    \vspace{0.05in}
    \centering
    \includegraphics[width = 0.7 \linewidth]{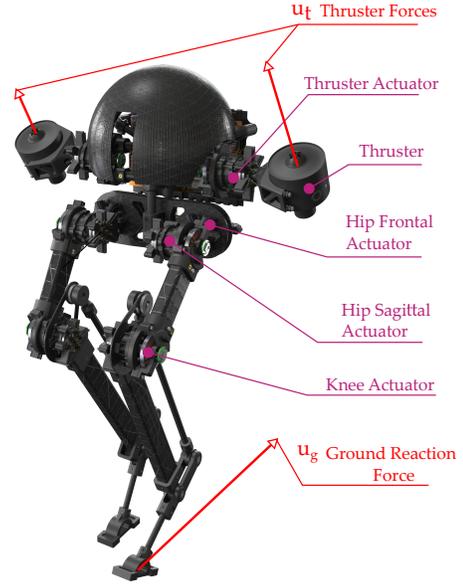}
    \caption{Illustration of a concept design for \textit{Harpy}, a thruster-assisted bipedal robot designed by the authors to study robust, efficient and agile legged robotics.}
    \label{fig:midget}
    \vspace{-0.05in}
\end{figure}

In this paper, we will report our efforts in designing closed-loop feedback for the thruster-assisted walking of a legged system called \textit{Harpy} (shown in Fig.~\ref{fig:midget}), currently its hardware being developed at Northeastern University. This biped is equipped with a total of eight actuators, and a pair of coaxial thrusters fixed to its torso. 
%
Our motivation stems from the merits of aerial and legged systems and we intend to integrate these merits in a single platform. Contrary to fixed- or rotary-wing aerial systems, legged robots cannot exhibit a fast mobility and fly over obstacles. A legged robot's capability to negotiate extremely bumpy terrains (e.g., semi-collapsed buildings in the aftermath of an earthquake) is very limited in that, for example, the height of terrain bumps should not exceed the size of the legs \cite{park2012finite,buss2014preliminary,park2012switching}. However, a legged robot maintains a superior energetic efficiency of locomotion because its overall body weight is supported by the legs, can safely operate inside buildings and has no sharp, rotating blades to cause severe laceration injuries to humans. 

Thruster-assisted legged locomotion has not been explored previously except to a limited extent in a few examples that only considered the hardware-related challenges \cite{picardi2019morphologically}. These robots potentially can offer rich and challenging dynamics and control problems. The overactuation and control allocation problems led by the coexistence of thrusters and joint actuators not only can provide opportunities to study interesting control ideas, but also, from a dynamical behavior standpoint, can permit studying unexplored behaviors such as walking under buoyancy phenomena \cite{kojio2016walking}. Also, studying multi-modal systems that can switch from one mode to another in order to overcome the demanding mobility objectives in unstructured environments is a rather new research problem and potentially can result in interesting machine-learning and optimization-based motion planning problems \cite{araki2017multi}. 
%

From a feedback design standpoint, the challenge of simultaneously providing asymptotic stability and gait feasibility constraints satisfaction in legged systems have been extensively addressed \cite{westervelt2007feedback}. For instance, the method of Hybrid Zero Dynamics (HZD) has provided a rigorous model-based approach to assign attributes such as efficiency of locomotion in an off-line fashion. Other attempts entail optimization-based, nonlinear approaches to secure safety and performance of legged locomotion \cite{CLFQP,7803333,7041347,ramezani2014performance,ramezani2013feedback,grizzle2013progress}.

Thrusters can result in unparalleled capabilities. For instance, gait trajectory planning (or re-planning), control and unilateral contact force regulation can be treated significantly differently as we have shown previously \cite{dangol2020performance,de2020thruster,dangol2020towards} and will further discuss new details in this paper. That said, real-time gait trajectory design in legged robots has been widely studied and the application of optimization-based methods is very common \cite{hereid20163d}. In general, in these paradigms, an optimization-based controller adjusts the gait parameters throughout the whole gait cycle such that not only the robot's posture is adjusted to accommodate the unplanned posture adjustments but also the joints position, velocity and acceleration are modified to avoid slipping into infeasible scenarios, e.g., the violation of contact forces. What makes these methods further cumbersome is that they are widely defined based on Whole Body Control (WBC) which can lead to computationally expensive algorithms \cite{sentis2006whole}. 

These problems are widely known to suffer from curse of dimensionality and other popular paradigms such as Approximate Dynamic Programming (ADP) \cite{powell2007approximate}, Reinforcement Learning (RL) \cite{sutton2018reinforcement}, decoupled approaches to design control for nonlinear stochastic systems \cite{rafieisakhaei2017near}, etc., can potentially remedy the challenges. However, these approaches are far from providing any practical solutions to the problem in hand and they are shown to be only effective on simpler practical robots mainly those that can only demonstrate quasi-static gaits. 

We will capitalize on the thrusters action in Harpy and will show that one can limit the use of costly optimization-based schemes by directly regulating contact forces. We will resolve gait parameters and re-plan them during the whole Single Support (SS) phase, which is the longest phase in a gait cycle, by only assuming well-tuned supervisory controllers found in \cite{sontag1983lyapunov, 371031, bhat1998continuous} and by focusing on fine-tuning the joints desired trajectories to satisfy unilateral contact force constraints. To do this, we will devise intermediary filters based on the celebrated idea of Explicit Reference Governors (ERG) \cite{garone2015explicit,411031,bemporad1998reference,gilbert2002nonlinear}. ERGs relied on provable Lyapunov stability properties can perform the motion planning problem in the state space in a much faster way than widely used optimization-based methods. That said, these ERG-based gait modifications and impact events (i.e., impulsive effects) can lead to severe deviations from the desired periodic orbits and standard legged robots cannot sustain these perturbations. Previously, we demonstrated that the thrusters can be leveraged to enforce hybrid invariance in a robust fashion by applying predictive schemes within the Double Support (DS) phase \cite{dangol2020towards}. Last, we also will show that thrusters can be leveraged to stabilize frontal dynamics and permit jumping over large obstacles which is an unusual capability not reported before. 

This paper is outlined as follows: the dynamic modeling for Harpy and the reduced-order models which will be used in the numerical simulation and controller design, the discussion on thruster assisted locomotion, numerical simulation discussions, and then followed by the concluding remarks.

\section{Dynamic Modeling of Harpy}
\label{sec:modeling}

\begin{figure}[t]
    \vspace{0.05in}
    \centering
    \includegraphics[width=0.6\linewidth]{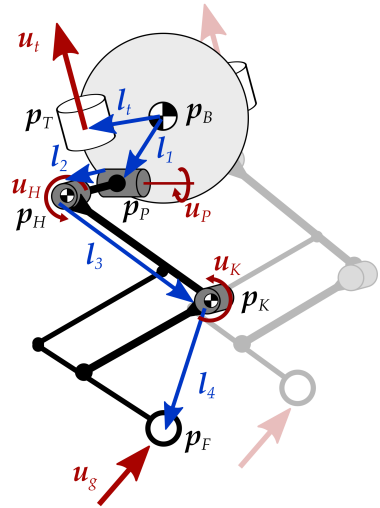}
    \caption{Shows the joint movements, key positions ($\bm p_i$) and dimensions ($\bm l_i$) of Harpy. The non-conservative forces and torques acting on the system are denoted by $\bm u_i$.}
    \label{fig:dof}
    \vspace{-0.05in}
\end{figure}

This section outlines the dynamics formulation of the robot which is used in the numerical simulation in Section \ref{sec:simulation}, in addition to the reduced order models which are used in the controller design. Fig. \ref{fig:dof} shows the kinematic configuration of Harpy which listed the center of mass (CoM) positions of the dynamic components, joint actuation torques, and thruster torques. The system model has a combined total of 12 degrees-of-freedoms (DoFs): 6 for the body and 3 on each leg. Due to the symmetry, the left and right side of the robot follow a similar derivations so only the general derivations are provided in this section.

\subsection{Euler-Lagrange Formalism}

The Harpy equations of motion are derived using Euler-Lagrangian dynamics formulation. In order to simplify the system, each linkages are assumed to be massless and the mass are concentrated at the body and the joint motors. Consequently, the lower leg kinematic chain is considered to be massless which significantly simplifies the system. The three leg joints are labeled as the hip frontal (pelvis $P$), hip sagittal (hip $H$) and knee sagittal (knee $K$), as illustrated in Fig. \ref{fig:dof}. The thrusters are also considered to be massless and capable of providing forces in any directions to simplify the problem. 


Let $\gamma_h$ be the frontal hip angle while $\phi_h$ and $\phi_k$ be the sagittal hip and knee angles respectively. Let the superscript $\{B,P,H,K\}$ represent the frame of reference about the body, pelvis, hip, and knee while the inertial frame is represented without the superscript. Let $R_B$ be the rotation matrix from the body frame to the inertial frame (i.e. $\bm x = R_B\, \bm x^B$). The pelvis motor mass is added to the body mass. Then the positions of the hip and knee CoM are defined using kinematic equations: 
\begin{equation}
\begin{gathered}
    \bm{p}_P = \bm{p}_{B} + R_{B}\, \bm{l}_{1}^{B}, \qquad
    \bm{p}_H = \bm{p}_{P} + R_{B}\,R_x(\gamma_h)\, \bm{l}_{2}^{P} \\
    \bm{p}_K = \bm{p}_{H} + R_{B}\,R_x(\gamma_h)\,R_y(\phi_h) \bm{l}_{3}^{H},
\end{gathered}
\label{eq:pos_com}
\end{equation}
where $R_x$ and $R_y$ are the rotation matrices about the $x$ and $y$ axis respectively, $\bm l$ is the length vectors representing the conformation of Harpy which are constant in their respective local frame of reference. 
The foot and thruster positions are defined as: 
\begin{equation}
\begin{gathered}
    \bm{p}_F = \bm{p}_{K} + R_{B}\,R_x(\gamma_h)\,R_y(\phi_h)\,R_y(\phi_k)\, \bm{l}_{4}^{K} \\
    \bm{p}_T = \bm{p}_{B} + R_{B}\, \bm{l}_{t}^{B}
\end{gathered}
\label{eq:pos_other}
\end{equation}
where the length vector from the knee to the foot is $\bm l_4^K = [-l_{4a}\cos{\phi_k}, 0, -( l_{4b} + l_{4a}\sin{\phi_k})]^\top$ which is the kinematic solution to the parallel linkage mechanism of the lower leg. Let $\bm \omega_B$ be the angular velocity of the body. Then the angular velocities of the hip and knee are defined as: $\bm \omega_H^B = [\dot{\gamma}_h,0,0]^\top + \bm \omega_B^B$ and $\bm \omega_K^H = [0,\dot{\phi}_h,0]^\top + \bm \omega_H^H$.

Finally, the energy of the system for the Lagrangian dynamics formulation are defined as follows:
\begin{equation}
\begin{aligned}
    K &= \tfrac{1}{2} \textstyle \sum_{i \in \mathcal{F}} \left( 
        m_i\,\bm p_i^\top\, \bm p_i + 
        \bm \omega_i^{i \top} \, \hat I_i \, \bm \omega_i^i \right) \\
    V &= - \textstyle \sum_{i \in \mathcal{F}} \left( 
        m_i\,\bm p_i^\top\, [0,0,-g]^\top \right),
\end{aligned}
\label{eq:energy}
\end{equation}
where $\mathcal{F} = \{B,H_L,K_L,H_R,K_R\}$ are the relevant frame of references and mass components (body, hip and knee of each side), and the subscripts $L$ and $R$ represent the left and right side of the robot. Furthermore, $\hat I_i$ is the inertia about its local frame, and $g$ is the gravitational constant. This forms the Lagrangian of the system $L = K-V$ which is used to derive the system's Euler-Lagrangian equations of motion. The dynamics of the body angular velocity is derived using the modified Lagrangian for rotation in SO(3) to avoid using Euler angles and the potential gimbal lock associated with them. This results in the following equations of motion following Hamilton's principle of least action:
\begin{equation}
\begin{gathered}
    \tfrac{d}{dt}\left( \tfrac{\partial L}{\partial \bm \omega_B^B}  \right) + 
    \bm \omega_B^B \times \tfrac{\partial L}{\partial \bm \omega_B^B} + 
    \textstyle \sum_{j=1}^{3} \bm r_{Bj} \times \tfrac{\partial L}{\partial \bm r_{Bj}} = \bm u_1 \\
    \tfrac{d}{dt}\left( \tfrac{\partial L}{\partial \dot {\bm q}}  \right) - 
    \tfrac{\partial L}{\partial \bm q} = \bm u_2, \qquad 
    \tfrac{d}{dt} R_B = R_B\, [\bm \omega_B^B]_\times,
\end{gathered}
\label{eq:eom_eulerlagrange}
\end{equation}
where $[\, \cdot \, ]_\times$ is the skew operator, $R_B^\top = [\bm r_{B1}, \bm r_{B2}, \bm r_{B3}]$, $\bm q = [\bm p_B^\top, \gamma_{h_L}, \gamma_{h_R}, \phi_{h_L}, \phi_{h_R}]^\top$ is the dynamical system states other than $(R_B,\bm \omega^B_B)$, and $\bm u$ is the generalized forces. The knee sagittal angle $\phi_k$ which is not associated with any mass is updated using the knee joint acceleration input $\bm u_k = [\ddot{\phi}_{k_L}, \ddot{\phi}_{k_R}]^\top$. Then the system acceleration can be derived as follows:
\begin{equation}
\begin{gathered}
    M \bm a + \bm h = B_j\, \bm u_j + B_t\, \bm u_t + B_g\, \bm u_g
\end{gathered}
\label{eq:eom_accel}
\end{equation}
where $\bm a = [ \dot{\bm \omega}_B^{B\top}, \ddot{\bm q}^\top, \ddot{\phi}_{k_L}, \ddot{\phi}_{k_R}]^\top$, $\bm u_t$ is the thruster force, $\bm u_j = [u_{P_L}, u_{P_R}, u_{H_L}, u_{H_R}, \bm u_k^\top]^\top$ is the joint actuation,  and $\bm u_g$ is the ground reaction forces (GRFs). The variables $M$, $\bm h$, $B_t$, and $B_g$ are a function of the full system states: 
\begin{equation}
    \bm x = [\bm r_{B}^\top, \bm q^\top, \phi_{K_L}, \phi_{K_R}, \bm \omega_B^{B \top}, \dot{\bm q}^\top, \dot{\phi}_{K_L}, \dot{\phi}_{K_R}]^\top,
\label{eq:states}
\end{equation}
where the vector $\bm r_B$ contains the elements of $R_B$. Using $B_j = [0_{6 \times 6}, I_{6 \times 6}]$ allows $\bm u_j$ to actuate the joint angles directly. Let $\bm v = [\bm \omega_B^{B\top}, \dot{\bm q}^\top]^\top$ be the velocity of the generalized coordinates, then $B_t$ and $B_g$ can be defined using the virtual displacement from the velocity as follows:
\begin{equation}
\begin{aligned}
    B_t = \begin{bmatrix}
        \begin{pmatrix}
        \partial \dot{\bm p}_{T_L} / \partial \bm v \\
        \partial \dot{\bm p}_{T_R} / \partial \bm v
        \end{pmatrix}^\top
        \\
        0_{2 \times 6}
    \end{bmatrix}, \quad
    B_g = \begin{bmatrix}
        \begin{pmatrix}
        \partial \dot{\bm p}_{F_L} / \partial \bm v \\
        \partial \dot{\bm p}_{F_R} / \partial \bm v
        \end{pmatrix}^\top
        \\
        0_{2 \times 6}
    \end{bmatrix}.
\end{aligned}
\label{eq:generalized_forces}
\end{equation}
The vector $\bm u_t = [\bm u_{t_L}^\top, \bm u_{t_R}^\top]^\top$ is formed from the left and thruster forces $\bm u_{t_L}$ and $\bm u_{t_R}$, respectively. 

The GRF is modeled using the unilateral compliant ground model with undamped rebound while the friction is modeled using the Stribeck friction model, defined as follows:
\begin{equation}
\begin{aligned}
    u_{g,z} =& -k_{g,p}\, p_{F,z} - k_{g,d}\, \dot p_{F,z} \\
    u_{g,x} =& -\left(\mu_c + (\mu_s - \mu_c)\, \mathrm{exp}\left(-\tfrac{|\dot p_{F,x}|^2}{v_s^2}\right) \right) f_z\, \mathrm{sgn}(\dot p_{F,x}) \\ 
    & - \mu_v\,\dot p_{F,x},
\end{aligned}
\label{eq:ground_model}
\end{equation}
where $p_{F,x}$ and $p_{F,z}$ are the $x$ and $z$ components of the inertial foot position, $k_{g,p}$ and $k_{g,d}$ are the spring and damping model for the ground, $\mu_c$, $\mu_s$, and $\mu_v$ are the Coulomb, static, and viscous friction coefficient respectively, and $v_s$ is the Stribeck velocity. $k_{g,d} = 0$ if $\dot p_{F,z} > 0$ for the undamped rebound model and the friction in the $y$ direction follows a similar derivation to $u_{g,x}$. Then the ground force model $\bm u_g$ is defined as follows:
\begin{equation}
\begin{aligned}
    \bm u_g = [\bm u_{g_L}^\top\, H(-p_{F_L,z}),\, \bm u_{g_R}^\top\, H(-p_{F_R,z})]^\top,
\end{aligned}
\label{eq:ground_forces}
\end{equation}
where $H(x)$ is the heaviside function, while $\bm u_{g_L}$ and $\bm u_{g_R}$ are the left and right ground forces which are formed using their respective $u_{g,x}$, $u_{g,y}$, and $u_{g,z}$. Finally, the full system equation of motion can be derived using \eqref{eq:eom_eulerlagrange} to \eqref{eq:ground_forces} to form $\dot{\bm x} = \bm f(\bm x, \bm u_j, \bm u_t, \bm u_g)$.

\subsection{Reduced-Order Models}

The following reduced-order models are used: a variable length inverted pendulum (VLIP) model and a two-body pendulum model, as illustrated in Fig. \ref{fig:reduced_order_models}. The VLIP model is used to describe walking with an ERG, while the two-body pendulum model is used to describe the flight phase with a Model Predictive Control (MPC) to track trajectory and regulate the appropriate leg postures at the time of landing.

\subsubsection{Variable-Length Inverted Pendulum (VLIP) Model} 

As shown in Fig. \ref{subfig:vlip}, the model is described simply using the inverted pendulum model where the length of $r$ can be adjusted through the change in leg conformation. The center of pressure, $\bm c$, is defined as the weighted average position of the foot, $\bm c = \lambda_L\, \bm p_{F_L} + \lambda_R\, \bm p_{F_R}$, where $\lambda_i = u_{g_i,z} / (u_{g_L,z} + u_{g_R,z})$, $i \in \{L,R\}$. Harpy is modeled using a point foot, so $\bm c$ is equal to the stance foot position during the SS phase. The VLIP model without thrusters is underactuated, but the addition of thrusters makes the system fully actuated and enables it to do trajectory tracking. 

The dynamic model is derived as follows:
\begin{equation}
\begin{gathered}
    m \ddot{\bm p}_B = m \bm g + \bm u_{t,c} + J_s^\top \bm \lambda\\ 
\end{gathered}
\label{eq:model_vlip}
\end{equation}
where $m$ is the mass of the pendulum which in this case is the total mass of the system, and $\bm u_{t,c}$ is the thruster forces about the body CoM. The constraint force $J_s^\top \bm \lambda$ is setup to keep the leg length $r$ equal to the leg conformation using the following constraint equation:
\begin{equation}
\begin{gathered}
    J_s\, (\ddot{\bm p}_B - \ddot {\bm c}) = u_r, \qquad   
    J_s = (\bm p_B - \bm c)^\top,
\end{gathered}
\label{eq:model_vlip_constraint}
\end{equation}
which is designed to keep the leg length second derivative equal to $u_r$. This constraint force also forms the GRF as long as the friction cone constraint is satisfied. Assuming no slip ($\ddot{\bm c} = 0$), then the inputs to the system are $u_r$ which controls the body position about the vector $\bm r = \bm p_B - \bm c$ by adjusting the leg length, and the thrusters $\bm u_t$ which controls the remaining DoFs. 

\subsubsection{Two-body Pendulum Model}

\begin{figure}[t]
    \vspace{0.05in}
    \centering
    \subfloat[\label{subfig:vlip}]{%
       \includegraphics[width=0.45\linewidth]{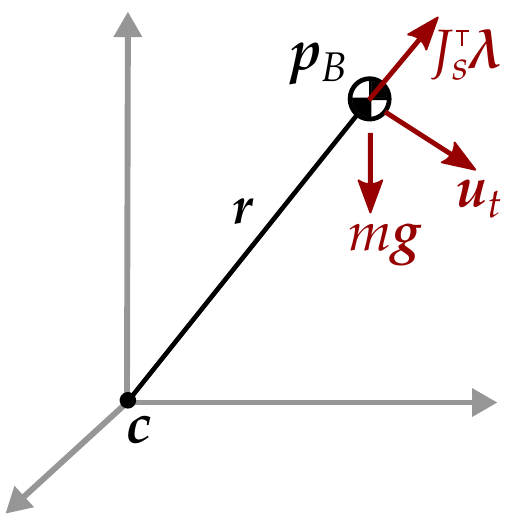}}
    \hfill
    \subfloat[\label{subfig:twobody}]{%
       \includegraphics[width=0.45\linewidth]{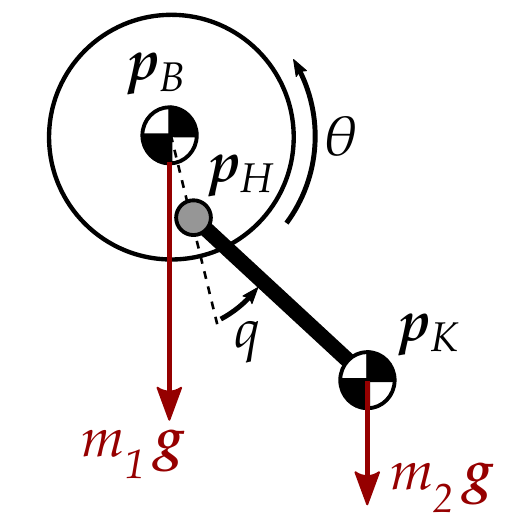}}
    \caption{Illustrates two reduced-order models, a variable-length inverted pendulum (VLIP) and a two-body pendulum model, which are used to describe the dominant dynamic response of the system during the walking and ballistic motions, respectively.}
    \label{fig:reduced_order_models}
    \vspace{-0.05in}
\end{figure}

As shown in Fig. \ref{subfig:twobody}, the model is described as a planar double pendulum in the $x$-$z$ plane where the mass is concentrated in the body CoM and knee, $m_1$ and $m_2$ respectively. Here, the mass of the pelvis motors are combined into the $m_1$ while the mass of both hips are combined into $m_2$ to ensure a similar kinematics behavior. 
%
The control action is defined as $\bm{u}_{dp} = [u_x, u_z, u_h]^\top$ where $u_x$ and $u_z$ are the sum of thruster forces in the $x$ and $z$ directions, respectively, and $u_h$ is the sagittal hip motor torque. The model uses the following states $\bm{q}_{dp} = [p_{B,x}, p_{B,z}, q, \theta]^\top$, where $q$ is the hip joint angle and $\theta$ is the body absolute pitch angle. Then the system acceleration can be derived as follows:
\begin{equation}
\begin{gathered}
    M_{dp} \ddot{\bm q}_{dp} + \bm h_{dp} = B_{dp}\bm u_{dp},
\end{gathered}
\label{eq:eom_accel_dp}
\end{equation}
where $B_{dp} = [ I_{3 \times 3},0_{1\times3}]$, allows the direct control signal to the system states except $\theta$.

\section{Control of Thruster-Assisted Locomotion}

This section discusses and outlines the applications of thrusters in our robot, particularly to stabilizes the frontal dynamics, apply the ERG framework to regulate GRFs, and to perform ballistic motions to avoid obstacles.

\subsection{Frontal Dynamics Stabilization}

In order to show the application of thruster-assisted walking in our robot, we use a walking gait designed for a planar bipedal robot. The absence of the frontal dynamics means that this gait is unstable for 3D walking. Hence the thrusters are used to stabilize the robot's roll and yaw to achieve a stable walking gait in the full 3D system.

The gait is designed by constraining the system dynamics in the $x$-$z$ plane using 4th order Bezier polynomials to define the feet-end positions relative to the hip in the robot body frame ($\bm p_F^B - \bm p_P^B$). Let $\bm P_i$, $i \in \{0,\dots,4\}$ be the Bezier polynomial parameters. This 4th order polynomial can be constrained to have zero initial and final gait velocity by setting $\bm P_0 = \bm P_1$ and $\bm P_3 = \bm P_4$. Then the free parameters $\bm P_0$, $\bm P_2$, and $\bm P_4$ define the initial, middle, and final positions of the gait. Each swing and stance feet-end have a constant Bezier curve parameters to form the 2D open-loop walking gait. The following parameters defined in the $x$-$z$ plane are used:
$\bm P_{0,sw} = [-0.21, -0.60]^\top$m , $\bm P_{2,sw} = [-0.20, -0.50]^\top$m, and $\bm P_{4,sw} = [0.10, -0.60]^\top$m for the swing foot, and $\bm P_{0,st} = [0.10, -0.60]^\top$m , $\bm P_{2,st} = [0.01, -0.63]^\top$m, and $\bm P_{4,st} = [-0.21, -0.60]^\top$m for the stance foot.
These Bezier parameters define the feet positions and then the joint angles are found simply by resolving the corresponding inverse kinematics problem. Finally, these joint trajectories are tracked using an asymptotically stable controller \cite{khalil2002nonlinear}.

The thrusters are used to stabilize the roll and yaw motion of the robot using the following controller
\begin{equation}
    \bm u_{t_L,F} = [u_{yaw},0,u_{roll}]^\top, \qquad
    \bm u_{t_R,F} = -\bm u_{t_L,F},
\label{eq:thruster_frontal}
\end{equation}
where $\bm u_{t_L,F}$ and $\bm u_{t_R,F}$ are the left and right thruster force components for frontal dynamics stabilization, while $u_{roll}$ and $u_{yaw}$ are simple PD controllers to track zero roll and yaw reference angles. This controller is sufficient to stabilize the frontal dynamics and the robot's heading even when using a gait designed for a 2D bipedal robot. During walking, the combined thruster forces are formed by combining $\bm u_{t,c}$ in \eqref{eq:model_vlip} and \eqref{eq:thruster_frontal}, as follows
\begin{equation}
    \bm u_t = [\bm u_{t,c}^\top, \bm u_{t,c}^\top]^\top/2 + [\bm u_{t_L,F}^\top, \bm u_{t_R,F}^\top]^\top.
\label{eq:thruster_walking}
\end{equation}

\subsection{Thruster-Assisted Enforcement of GRF Constraints Using RG-based Methods}

The ERG framework is utilized to enforce the friction pyramid constraint of the robot by manipulating the applied reference to the controller, which is useful to avoid using optimization or nonlinear MPC framework to enforce constraints on the harder-to-model GRFs. The VLIP model in \eqref{eq:model_vlip} can be fully actuated due to the addition of thrusters, which enables us to utilize a more advanced controller such as ERG. 

\begin{figure}[t]
    \vspace{0.05in}
    \centering
    \includegraphics[width=0.7\linewidth]{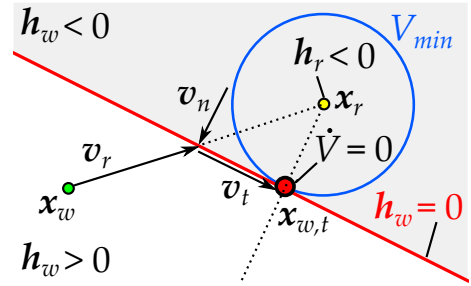}
    \caption{Shows how an Explicit Reference Governor (ERG) can manipulate the applied reference states $\bm x_w$ to be as close as possible to the desired reference $\bm x_r$ while the constraint equations $\bm h_w \geq 0$ remain feasible.}
    \label{fig:erg}
    \vspace{-0.05in}
\end{figure}

The ERG manipulates the applied reference ($\bm x_w$) to avoid violating the constraint equation $\bm h_w(\bm x, \bm x_w) \geq 0$ while also be as close as possible to the desired reference ($\bm x_r$), as illustrated in Fig. \ref{fig:erg}. Consider the Lyapunov equation $V = (\bm x_r - \bm x_w)^\top P (\bm x_r - \bm x_w)$. $\bm x_w$ is updated through the update law:
\begin{equation}
    \dot{\bm x}_w = \bm v_r + \bm v_t + \bm v_n,
\label{eq:erg_update}
\end{equation}
where $\bm v_r$ drives $\bm x_w$ directly to $\bm x_r$, while $\bm v_t$ and $\bm v_n$ drives $\bm x_w$ along the surface and into the constraint equation boundary $\bm h_w =  0$, respectively. The objective of this ERG algorithm is to drive $\bm x_w$ to the state $\bm x_{w,t}$ which is the minimum energy solution $V_{min}$ that satisfies the constraint $\bm h_w \geq 0$.


Let $\bm h_r(\bm x, \bm x_r) = J_r \bm x_r + \bm d_r \geq \bm 0$ be the constraint equation using the desired reference $\bm x_r$, and $C_r$ be the rowspace of the violated constraints of $\bm{h}_r$. Define $N_r = \mathrm{null}(C_r) = [\bm{n}_1, \dots, \bm{n}_{n}]$ where $n$ is the size of the nullspace. Additionally, let $\bm r_k$ be the $k$'th row of $J_r$. Then the following update law is used for $\bm v_r$, $\bm v_t$, and $\bm v_n$:
\begin{equation}
\begin{gathered}
    \bm v_r = \hat \alpha_r\, (\bm x_r - \bm x_w), \qquad
    \bm v_n = \hat \alpha_n\,\bm r_k\,\bm r_k^\top\, (\bm x_r - \bm x_w) \\
    \bm v_t = \textstyle \sum^n_{k=1} \hat{\alpha}_t\, \bm{n}_k\,\bm{n}_k^\top (\bm{x}_r - \bm{x}_w),
\end{gathered}
\label{eq:erg_update_v}
\end{equation}
where $\hat \alpha$ are scalars defined as follows:
\begin{equation}
\begin{aligned}
    \hat \alpha_r &= 
    \begin{cases}
     \alpha_r, & \text{if } \min(\bm h_w) \geq 0 \text{ or } \min(\bm h_r) \geq 0 \\
     0,      & \text{else} \\
    \end{cases} \\
    \hat \alpha_t &= 
    \begin{cases}
     \alpha_t, & \text{if } \min(\bm h_w) \geq 0 \text{ or } \min(\bm h_r) < 0 \\
     0,      & \text{else} \\
    \end{cases} \\
    \hat \alpha_n &= 
    \begin{cases}
     \alpha_n, & \text{if } \min(\bm h_w) \leq \min(\bm h_r) < 0 \\
     -\alpha_n, & \text{if } \min(\bm h_r) < \min(\bm h_w) < 0 \\
     0,      & \text{else,} \\
    \end{cases}
\end{aligned}
\label{eq:erg_update_alpha}
\end{equation}
where $\alpha$ is a positive scalar which determines the rate of convergence. 

Assuming $\dot{\bm x}_r = 0$ and using the update law defined from \eqref{eq:erg_update_v} and \eqref{eq:erg_update_alpha}, we will have $\dot{V} = - 2(\bm x_r - \bm x_w)^\top\, Q\, (\bm x_r - \bm x_w)$, with $Q = P ( \hat \alpha_r\, I + \textstyle \sum^n_{k=1} \hat \alpha_t\, \bm{n}_k \,\bm{n}_k^\top + \hat \alpha_n\,\bm r_k\,\bm r_k^\top)$. We have the gradient $\dot{V} = 0$ if $\min(\bm h_w) \leq 0$ and $\bm n_k \bot (\bm x_r - \bm x_w)$, while $\dot{V} < 0$ when $\min(\bm h_r) \geq 0$ or when $\min(\bm h_w) \geq 0$. In case both applied reference and target constraints equation are violated, we have $\dot V > 0$ which drives the $\bm x_w$ towards the constraint boundary. This allows the $\bm x_w$ to converge to $\bm x_{w,t}$ which is the minimum energy solution that satisfies $\bm{h}_w \geq 0$ as illustrated in Fig. \ref{fig:erg}. 

The GRF constraints for this robot can be derived from the reduced-order model constraint equation in \eqref{eq:model_vlip_constraint} by applying the ground pyramid constraint. We use the following constraints for the ERG:
\begin{equation}
    |u_{g,x}| \leq \mu_s u_{g,z}, \quad
    |u_{g,y}| \leq \mu_s u_{g,z}, \quad
    u_{g,z} \geq u_{z,min},
\label{eq:erg_constraints}
\end{equation}
where $[u_{g,x}, u_{g,y}, u_{g,z}]^\top = J_s^\top \bm \lambda$ is the ground reaction force model from \eqref{eq:model_vlip} and \eqref{eq:model_vlip_constraint}. This forms the ground friction pyramid constraint and the minimum ground normal force acting on the leg to ensure that the foot doesn't slip.

\subsection{MPC Design and Control of Robot's Ballistic Motion}

The thrusters can also be used to traverse rough terrain by simply flying over obstacles. However, it is not possible to stabilize pitch dynamics due to the thrusters position which is aligned about the body CoM sagittal axis. Therefore, the reduced-order model in \eqref{eq:eom_accel} is used to model the robot during the ballistic motion which is represented as a two-body pendulum system. Then an MPC framework is developed based on this model to regulate the feet-end positions and ensuring the proper landing configuration. 

The hip frontal and knee angles are setup to be constant during the ballistic maneuver and the feet are positioned in a neutral stance configuration within which legs are parallel to each other. An MPC with a prediction and control horizons $N_h = 10$ is developed using the following optimizer
\begin{equation}
\begin{gathered}
    \min_{\bm{u}_{dp}} \textstyle{\sum_{k=0}^{N_h} (\boldsymbol{e}^{k\top}\, W_{e}\, \boldsymbol{e}^k + \Delta \boldsymbol{u}_{dp}^{k\top} \,W_{u}\, \Delta \boldsymbol{u}_{dp}^k )} \Delta t \\
    \begin{aligned}
    \text{subject to} \qquad & & 
    \dot{\bm x}_{dp} &= f(\bm x_{dp},\bm u_{dp}) \\
    && q_{ref}^{k+1} - \theta^k + q_c &= 0 \\
    && \left|\boldsymbol{u}_{dp}\right| - \boldsymbol{u}_{dp,max } &\leq 0,
    \end{aligned}
\end{gathered}
\end{equation}
where $\bm e = [\bm p_{B}^\top - \bm p_{B,ref}^\top, q -  q_{ref}]^\top$ is the tracking error, $W_e$ and $W_u$ are the cost weighting matrices, and $q_c$ is a constant angle to determine the desired landing posture. $q_{ref}$ denotes the reference hip sagittal angle which is updated each time step relative to the body pitch angle $\theta$ and $q_c$. This reference can ensure the legs do not lag behind the body at the landing moment, and $q_c$ can be properly adjusted to achieve the desired landing posture. The references $\bm p_{B,ref}$ are designed as a ballistic motion for the body CoM as follows
\begin{equation}
    \bm p_{B,ref}^{k+1} = \bm p_{B,ref}^{k} + \Delta t\, [a, b \sin(2\pi t/1.5)]^\top
\end{equation}
where $\bm p_B$ denotes the body position in the $x$-$z$ plane, while $a$ and $b$ are some constants, and $\Delta t$ is the controller time step. Finally, the resulting $\bm u_{dp} = [u_x, u_z, u_h]$ is fed to the full model, where the thruster forces $u_x$ and $u_z$ are combined with the roll and yaw stabilization controller in \eqref{eq:thruster_frontal}. This forms the combined thruster forces
\begin{equation}
    \bm u_t = [u_x, 0, u_z, u_x,0, u_z]^\top / 2 + [\bm u_{t,l}^\top, \bm u_{t,r}^\top]^\top,
\label{eq:combined_thruster_force}
\end{equation}
which tracks the flight trajectory and stabilizes the robot's roll and heading angles.


\section{Simulation Results}
\label{sec:simulation}

\begin{figure*}
    \centering
    \subfloat[\label{subfig:animation}]{%
    \includegraphics[width=0.67\linewidth]{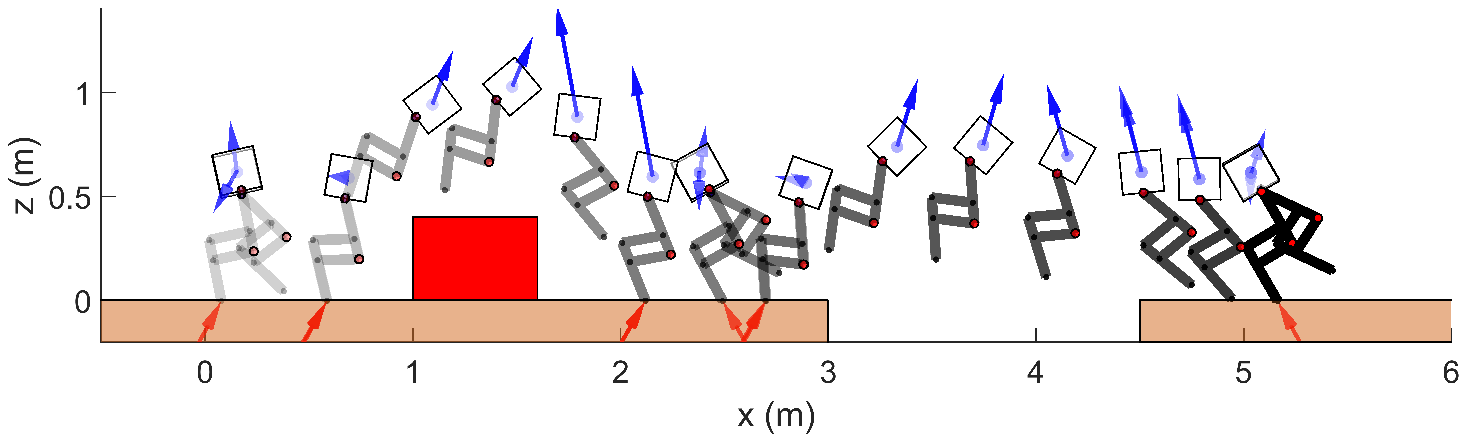}}
    \vspace{-0.1in}
    \centering
    \subfloat[\label{subfig:data}]{%
       \includegraphics[width=0.99\linewidth]{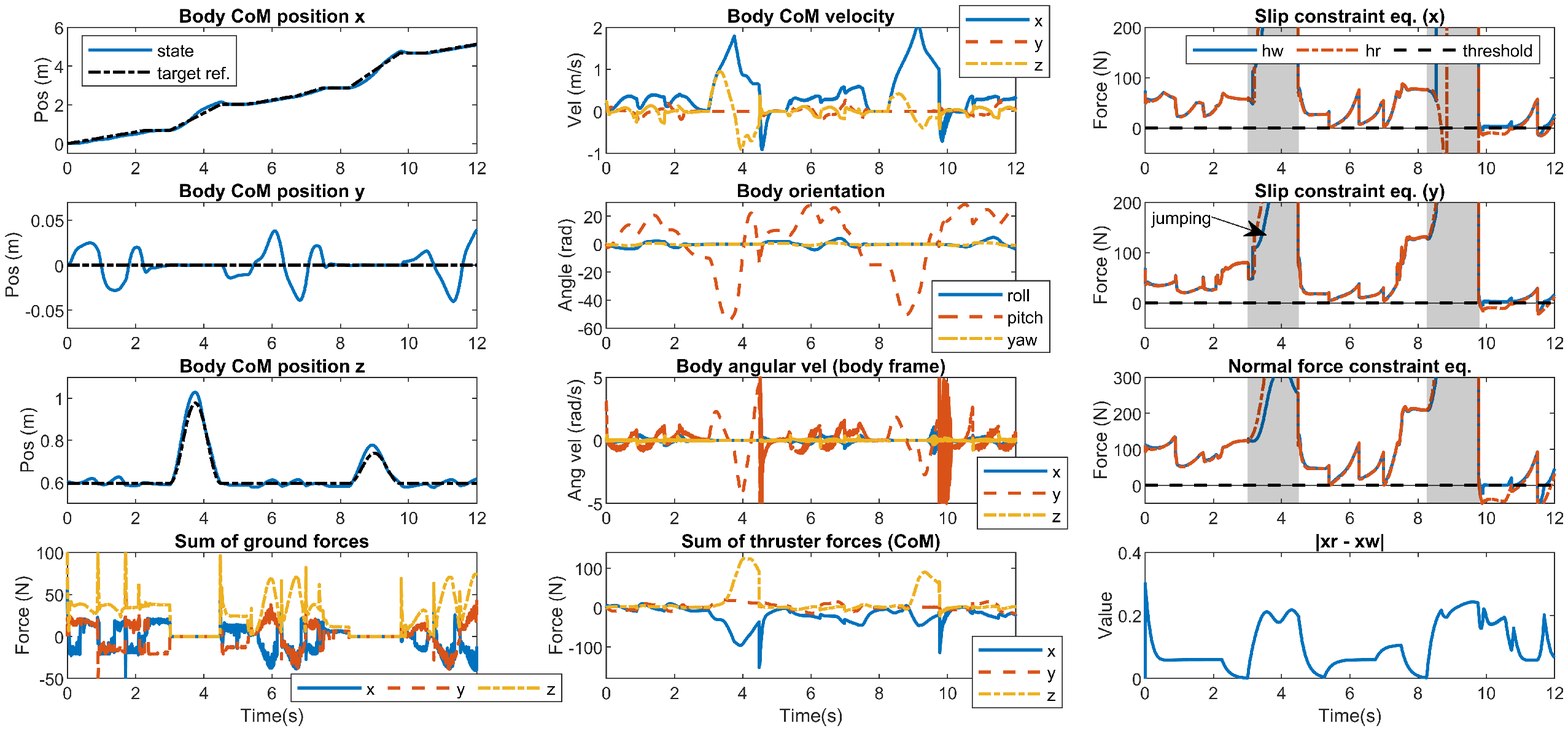}
       }
    \vspace{-0.1in}   
    \centering
    \subfloat[\label{subfig:data_zoomed}]{%
       \includegraphics[width=0.99\linewidth]{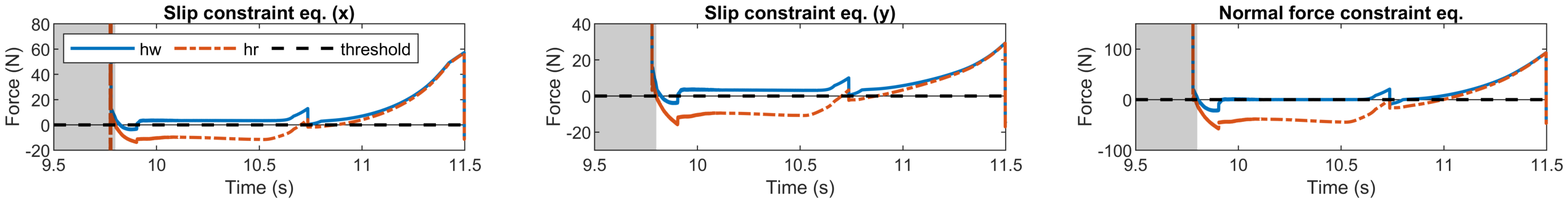}
       }
       
    \caption{Shows the simulation result for the walking and jumping maneuvers. (a) Illustrates the stick-diagram plot when the robot walks on flat ground and jumps over obstacles. (b) Depicts the states evolution, the thrusters action and ground contact forces. Note that the thrusters are employed to stabilize the frontal dynamics. The RG algorithm is disabled during the ballistic motion (grey area). (c) Close-up view of the constraints at $t = [9.5,11.5]$s where the applied references ($\bm x_w$) are manipulated such that the constraints ($\bm h_w(\bm x, \bm x_w) \geq 0$) remain feasible.}
    \label{fig:result}
    \vspace{-0.1in}
\end{figure*}

The optimization and numerical simulation are done in Matlab where we used interior-point algorithm and RK4 scheme, respectively. The MPC and the ERG filter are simulated using a zero-order hold at a frequency 20 times smaller than the numerical integration to better capture the behavior of the onboard computer in Harpy. 

\subsection{Simulation Specifications}

In this section, all units are in N, kg, m, s. The robot has the following dimensions: $\bm l_1 = [0, 0.1, -0.1]^\top$, $\bm l_2 = [0, 0.5, 0]^\top$, $\bm l_3 = [0, 0, -0.3]^\top$, and $\bm l_4 = [0, 0.1, 0]^\top$ for the left side. The right side simply has the $y$ axis component inverted. This gives the robot CoM height of around 0.6 m for standing and walking. The following mass and inertias are used: $m_B = 2$, $m_H = m_K = 0.5$, $I_B = 10^{-3}$, $I_H = I_K = 10^{-4}$. Finally, the following ground parameters are used: $\mu_s = 0.6$, $\mu_c = 0.54$, $\mu_v = 0.85$, $k_{g,p} = 8000$, and $k_{g,d} = 268$. The frequency of MPC controller applied in the two-body pendulum model is 100 Hz with a 10 steps prediction and control horizons. The entire simulation contains 16 gait cycles (12 s) using the following sequence: 3 walking steps, 1 quiet standing cycle, 2 jumping cycles, 4 walking steps, 1 quiet standing cycle, 2 jumping cycles, and 3 walking steps. The weighting matrices for the MPC are $W_e = \mathrm{diag}(5,20)$, $W_u = \mathrm{diag}(1/10,1/10,1/5)$, $q_c = 23^{\circ}$, and the parameters for the jumping trajectories are $a = 0.9$, $b = 0.8$ for the first jumping cycle, and $a = 1.2$, $b = 0.3$ for the second jump. The ERG is applied using $\alpha=5$ which provides a sufficient convergence rate.

\subsection{Simulation Results and Discussions}

The simulation results can be seen in Fig.~\ref{fig:result}, where Fig.~\ref{subfig:animation} shows the key frames of the simulated robot trajectory where it walks and jumps over obstacles. Figure~\ref{subfig:data} shows the data of the thruster forces, ground normal forces, and body states during the simulation. The walking gait designed for a 2D bipedal robot is stable when used in the full 3D system as the pitch and yaw angles are stabilized by the thruster actions. Additionally, the trajectory of the body positions match closely towards the desired trajectories throughout the walking and flight phases. The robot's pitch angle is uncontrollable as shown in Fig.~\ref{subfig:data} throughout the entire simulation. However, the MPC has successfully regulated the foot landing positions such that the foot is positioned below or in front of the body at the time of landing through the appropriate hip sagittal angles, as shown in Fig.~\ref{subfig:animation}. This allows a smooth transition between landing and walking which is the primary objective of using the MPC.


The ERG is used to regulate the ground friction forces by manipulating the applied state references during walking to prevent slips. Figure~\ref{subfig:data} shows the constraint equation values using the applied and target reference ($\bm h_w$ and $\bm h_r$ respectively), where the target reference trajectories satisfy the constraints except at around the time range of $t = [10,12]$s as shown in Fig.~\ref{subfig:data_zoomed}. Within this time range, the desired reference trajectory results in $\min(\bm h_r) < 0$ and the ERG has successfully manipulate the applied reference $\bm x_w$ such that the constraint equation $\bm h_w \geq 0$ is satisfied. The control output of the ERG is disabled during the jumping period (shaded gray in Fig.~\ref{subfig:data}) as the thruster components for position stabilization is handled by the MPC.

\section{Conclusions and Future Work}
\label{sec:conclusions}

The concept design and dynamics simulation of a thruster-assisted bipedal robot called Harpy is presented in this paper. The addition of the thrusters allows the robot to stabilize its frontal dynamics easily and be able to regulate the ground contact forces directly. To do this, a control algorithm based on Reference Governors (RGs) was proposed. In addition, the thrusters allowed multimodal locomotion where the robot can transition between walking and jumping over obstacles. Through simulations we showed that the integration of the thrusters can yield a dynamically robust biped. Future work includes the completion of Harpy. In addition, we will explore the cost of transport in multi-modal systems. Further investigations will be made in developing high fidelity models that incorporate realistic thruster dynamics which are more appropriate for the real-world implementation of closed-loop feedback on Harpy.


\bibliographystyle{IEEEtran}
\bibliography{reference}

\end{document}